\documentclass[a4paper,conference]{IEEEtran}
\IEEEoverridecommandlockouts 
\usepackage{amsmath,amssymb,amsfonts}
\usepackage{algorithm}
\usepackage{algpseudocode}
\usepackage{tabularx}
\usepackage{graphicx,,epstopdf}
\usepackage{graphics}
\usepackage{graphicx}
\usepackage{epstopdf}
\usepackage{subcaption}
\usepackage{textcomp}
\usepackage{xcolor}
\usepackage{blindtext}
\usepackage{multirow}
\usepackage{mathtools, nccmath}
\usepackage{hyperref}
\usepackage{cite}
\usepackage{upgreek}
\usepackage[a4paper,
            left=54pt,
            right=37pt,
            top=104pt,
            bottom=54pt]{geometry}

\newtheorem{assumption}{Assumption}
\newtheorem{remark}{Remark}

\makeatletter
\newcommand{\multiline}[1]{%
  \begin{tabularx}{\dimexpr\linewidth-\ALG@thistlm}[t]{@{}X@{}}
    #1
  \end{tabularx}
}
\makeatother

\newcommand{\argmin}{\mathop{\mathrm{argmin}}\limits} 

\def\BibTeX{{\rm B\kern-.05em{\sc i\kern-.025em b}\kern-.08em
    T\kern-.1667em\lower.7ex\hbox{E}\kern-.125emX}}

\title{\vspace{18pt}Distributed Control for 3D Inspection using Multi-UAV Systems}

\author{Angelos Zacharia, Savvas Papaioannou, Panayiotis Kolios and Christos Panayiotou 
\thanks{The authors are with the KIOS Research and Innovation Center of Excellence (KIOS CoE) and the Department of Electrical and Computer Engineering, University of Cyprus, Nicosia, 1678, Cyprus.\{zacharia.angelos, papaioannou.savvas, pkolios, christosp\}@ucy.ac.cy}}

\begin{document}
\maketitle

\begin{abstract}
Cooperative control of multi-UAV systems has attracted substantial research attention due to its significance in various application sectors such as emergency response, search and rescue missions, and critical infrastructure inspection. This paper proposes a distributed control algorithm to generate collision-free trajectories that drive the multi-UAV system to completely inspect a set of 3D points on the surface of an object of interest. The objective of the UAVs is to cooperatively inspect the object of interest in the minimum amount of time. Extensive numerical simulations for a team of quadrotor UAVs inspecting a real 3D structure illustrate the validity and effectiveness of the proposed approach.
\end{abstract}

	
\section{Introduction}
\label{introduction}
Unmanned Aerial Vehicles (UAVs) have become increasingly popular in recent years due to their adaptability and wide range of uses. UAVs equipped with complementary sensor payloads such as cameras, radars, and navigation systems (e.g. GPS) allow their use in various applications, such as surveillance \cite{kingston2016automated}, emergency response missions \cite{papaioannou2020cooperative}, security \cite{papaioannou2020cooperativeSecurity,souli2020horizonblock}, and inspection \cite{savva2021icarus}. The ability of UAVs to access hard-to-reach or dangerous areas is exploited to provide a cost-effective and efficient solution for several tasks.

One of the most prevalent applications of automated UAV-based systems is the inspection/coverage of an object of interest (e.g., collapsed buildings, critical infrastructures, sensitive facilities), which involves the process of determining the trajectory of a UAV to fully inspect/cover a specific area effectively. However, inspection/coverage planning for UAVs is a challenging problem due to the balancing of multiple objectives such as inspection quality and completeness, mission duration, and energy consumption. Inspection/coverage trajectory planning aims to ensure that each UAV agent provides detailed information about the object being inspected using technological equipment, such as a gimballed camera and/or LiDAR, while also not violating the UAV's dynamics, sensing constraints, and reducing the risk of collisions.

In the literature, numerous methodologies have been proposed to solve the well-established problem of inspection/coverage planning in a three-dimensional environment using either single-robot or multi-robot systems. The authors of \cite{englot2013three} extract stationary viewpoints for an underwater inspection robot that ensures complete coverage based on redundant roadmap and watchman route algorithms. Then, the traveling salesman problem (TSP) is solved to find a feasible path to inspect the hull of the ship. Similarly, in \cite{jing2016sampling}, a sampling-based view planning approach is proposed for a camera-equipped UAV to capture visual geometric data of objects of interest. In \cite{papaioannou2021towards,papaioannou20213d}, the 3D inspection/coverage planning is transformed into mixed integer quadratic programming, generating UAV trajectories capable of covering cuboid-like objects of interest. Our previous work in \cite{papaioannou2022uav} proposes a UAV-based receding horizon inspection planning control methodology that generates an optimal trajectory for inspecting crucial feature-points on the object's surface. For 3D surface inspection, the authors in \cite{zhu2021online} propose an online approach consisting of the optimal control waypoints extraction process and the generation of the continuous UAV's trajectory from these points. However, the primary limitation of the methodologies mentioned above is the number of robots used to solve the problem of 3D inspection/coverage planning. Optimization of a single-robot path does not significantly reduce inspection time compared to the use of a multi-robot system, which provides faster and more robust 3D inspection.

In recent related works, researchers have also investigated the problem of multi-UAV 3D inspection/coverage trajectory planning. For example, the work in \cite{mansouri2018cooperative} examines the cooperative inspection of a complex 3D structure using a UAV team, achieving complete coverage through the division of the infrastructure's surface and assigning resulting areas to each UAV. The problem is then formulated as a problem of multiple traveling salesman. The main drawback of this work is that the trajectories of UAVs are generated offline in a centralized manner. In \cite{ahmed2021distributed}, the authors proposed a trajectory planner for a team of multiple UAVs based on the particle swarm optimization approach, which finds optimal trajectories using distributed full coverage and a dynamic fitness function. Finally, in \cite{ivic2023multi}, the authors propose a heat-equation-driven area coverage methodology for visual inspection of complex 3D structures using a multi-UAV system. The algorithm produces collision-free trajectories and UAV camera orientations. The computational inefficiency of the proposed methods in \cite{ahmed2021distributed,ivic2023multi} is one of their main drawbacks.

Motivated by the discussion mentioned above, this paper presents a planning algorithm to achieve complete 3D inspection of an object of interest using a multi-UAV system. Specifically, the object of interest consists of a finite number of distinct 3D points on its surface that multiple quadrotor UAVs cooperatively inspect. It is assumed that all UAVs are identical, governed by the same dynamical model, and have similar sensing capabilities. Given these assumptions, a 3D inspection planning algorithm is proposed for a UAV team that determines each UAV's distributed control inputs, with the resulting trajectories to allow complete inspection of the object of interest. The inspection trajectories are computed online by each UAV using only local measurements and information shared by its neighbors. The paper's contributions are summarized as follows:

\begin{itemize}
    \item We propose an online trajectory planning algorithm that enables the multi-UAV system to fully inspect large-scale complex structures in 3D environment.
    \item We design a distributed control scheme for each UAV that uses only local measurements, such as the UAV's position and velocity, as well as its neighbors' positions.
\end{itemize}

The rest of the paper is organized as follows. In Section \ref{systemModel}, we develop the system model based on our modeling assumptions, while the problem addressed by this work is outlined in Section \ref{problemStatement}. In the sequel, Section \ref{proposedApproach} discusses the details of the proposed 3D inspection approach, and Section \ref{evaluation} evaluates the proposed methodology. Finally, Section \ref{conclusion} concludes the paper and discusses future work.

\section{System Model}
\label{systemModel}

\subsection{UAV Dynamics}
The quadrotor UAV navigates within a bounded and convex region $\mathcal{Q}\subset \mathbb{R}^3$. An earth-fixed coordinate frame $\mathbf{E} = \{x_{\mathbf{E}},y_{\mathbf{E}},z_{\mathbf{E}}\}$ is arbitrarily positioned in the 3D space, while a body-fixed coordinate frame $\mathbf{B} = \{x_{\mathbf{B}},y_{\mathbf{B}},z_{\mathbf{B}}\}$ is attached to the quadrotor UAV, with its origin coinciding with the center of mass of the agent. To simplify the complex dynamics of the UAV, we employ the feedback linearization control technique, as in \cite{hernandez2015trajectory}, which allows the use of linear controllers. Thus, we can model the UAV dynamics as a double-integrator with the following equations of motion:

\begin{equation}
    \begin{split}
        \dot{\mathbf{p}} &= \boldsymbol{\upsilon}\\
        \dot{\boldsymbol{\upsilon}} &= \boldsymbol{u}
    \end{split}
    \label{UAVdynamics}
\end{equation}
where the position of the UAV is denoted by $\mathbf{p} = [x,y,z]^T \in \mathbf{E}$, $\boldsymbol{\upsilon}\in \mathbf{E}$ represents the linear velocity vector, and $\boldsymbol{u} = [u_x,u_y,u_z]^T \in \mathbb{R}^{3\times 1}$ denotes the control input vector.

\subsection{Object 3D Representation}
\begin{figure*}[t]
     \begin{subfigure}[b]{0.24\textwidth}
        \includegraphics[clip, trim=2cm 15cm 9cm 2cm, scale = 0.6]{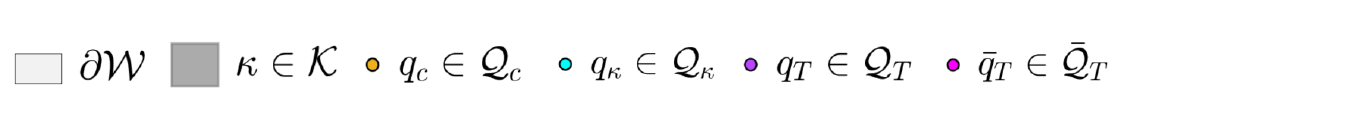}
    \end{subfigure}
    
    \centering
     \begin{subfigure}[b]{0.24\textwidth}
        \includegraphics[clip, trim=2cm 8cm 1cm 8cm, width=1.15\textwidth]{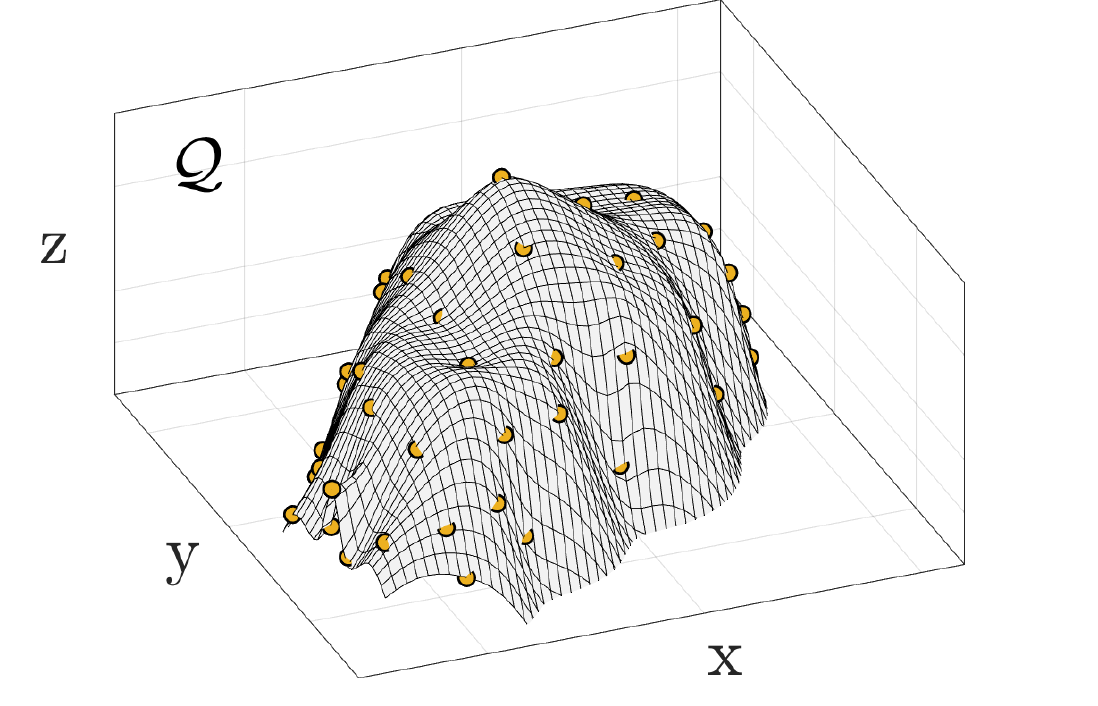}
        \caption{}
        \label{fig:point-cloud}
    \end{subfigure}
    \hfill
    \begin{subfigure}[b]{0.24\textwidth}
        \includegraphics[clip, trim=2cm 8cm 1cm 8cm, width=1.15\textwidth]{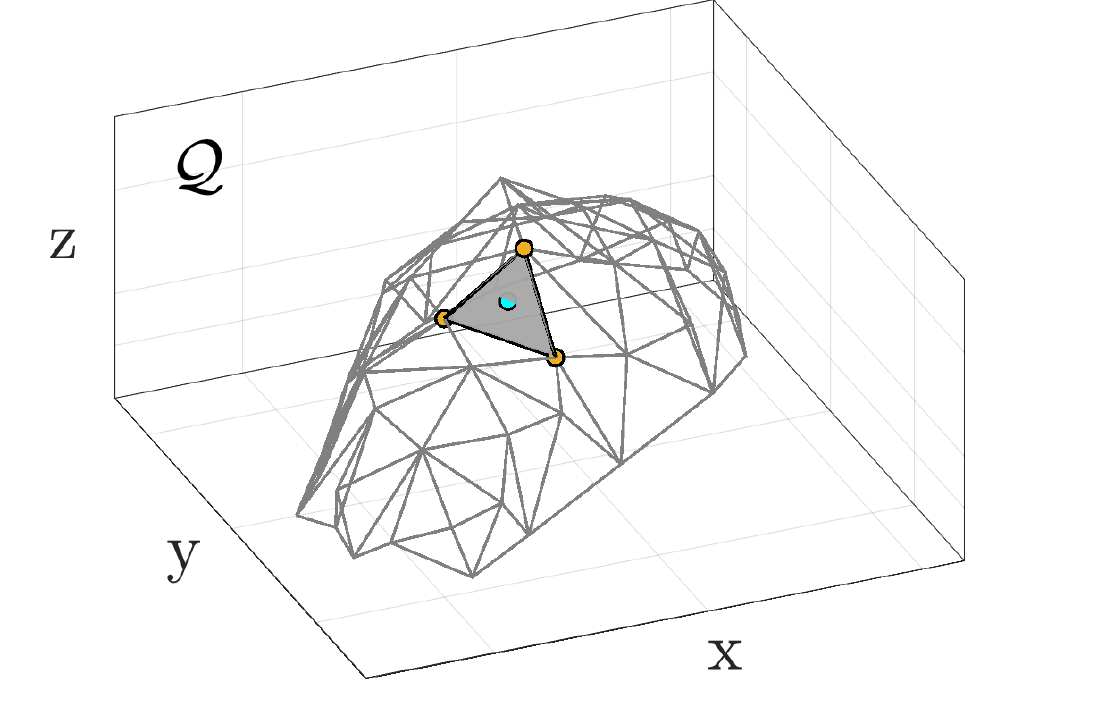}
        \caption{}
        \label{fig:triangulation}
    \end{subfigure}
    \hfill
    \begin{subfigure}[b]{0.24\textwidth}
        \includegraphics[clip, trim=2cm 8cm 1cm 8cm, width=1.15\textwidth]{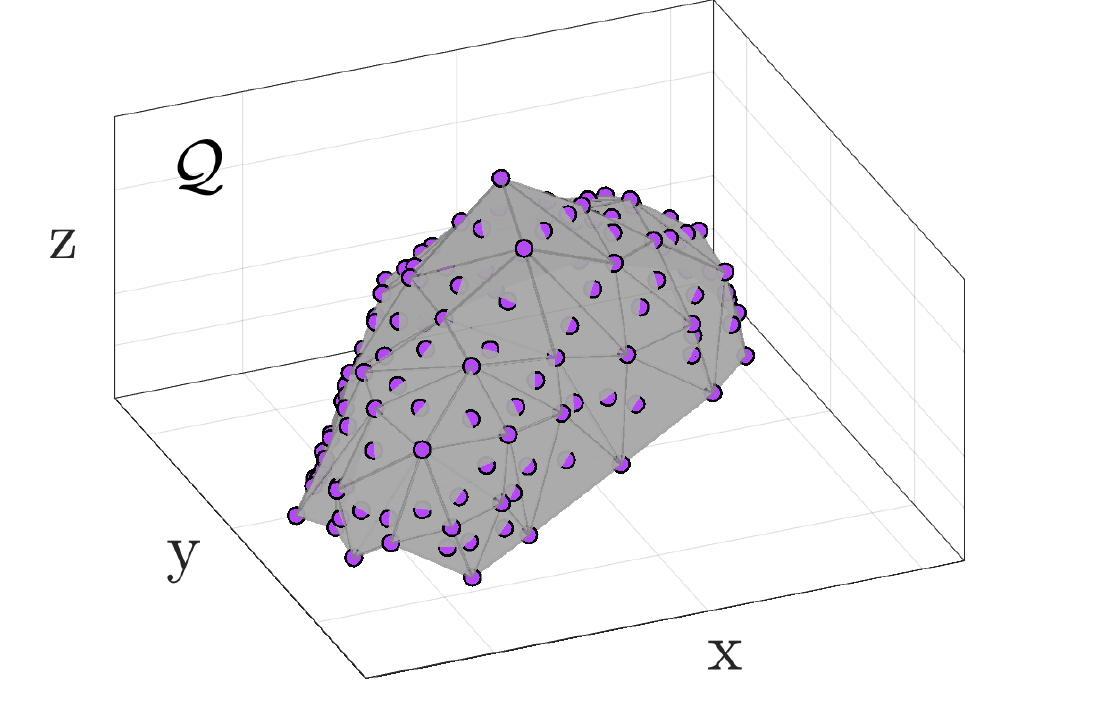}
        \caption{}
        \label{fig:targetPoints}
    \end{subfigure}
    \hfill
    \begin{subfigure}[b]{0.24\textwidth}
        \includegraphics[clip, trim=2cm 8cm 1cm 8cm, width=1.15\textwidth]{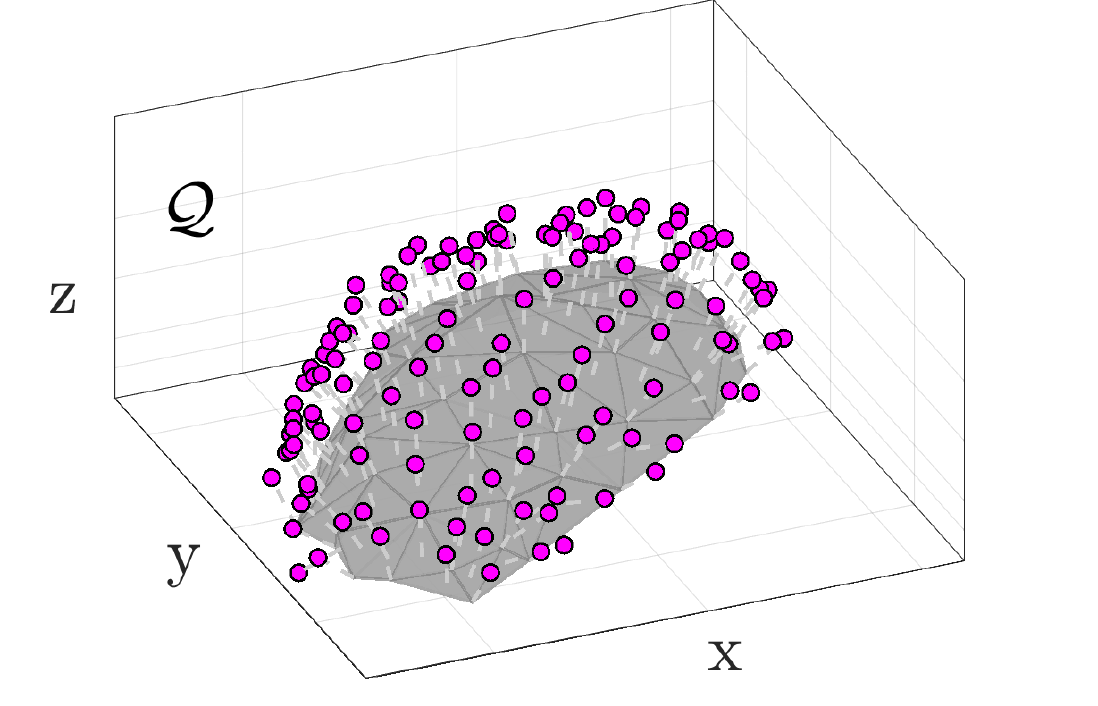}
        \caption{}
        \label{fig:projectedTargetPoints}
    \end{subfigure}
            
    \caption{(a) 3D point-cloud $\mathcal{Q}_c$ generation from the object's boundary $\partial\mathcal{W}$, (b) Triangle mesh $\mathcal{K}$ formed by Delaunay triangulation, (c) Target points to be inspected, and (d) Outward projection of target points}
    \label{fig:object}
\end{figure*}

To produce a 3D representation of an object of interest $\mathcal{W}\subset \mathcal{Q}$, several calibrated images must be collected during the 3D reconstruction process \cite{moons20103d}. A 3D point-cloud $\mathcal{Q}_c = \{q_c^l\}$, $l \in \{1,\ldots,|\mathcal{Q}_c|\}$, is derived from these images, with points $q_c \in \mathcal{Q}_c$ corresponding to the surface area on the boundary $\partial \mathcal{W}$ of the object, as shown in Fig. \ref{fig:point-cloud}. Using the Delaunay triangulation method \cite{wang2022restricted}, a triangle mesh $\mathcal{K}$ is formed, consisting of triangular facets $\kappa \in \mathcal{K}$, as depticted in Fig. \ref{fig:triangulation}. Furthermore, the center of each triangular facet $q_{\kappa} \in \mathcal{Q}_{\kappa}$ is computed and combined with the 3D point-cloud set $\mathcal{Q}_c$ to create the target set $\mathcal{Q}_T = {\mathcal{Q}_c \cup \mathcal{Q}_{\kappa}}$, as presented in Fig. \ref{fig:targetPoints}. The resulting target set is the set of points that the multi-UAV system has to cooperatively inspect.

\section{Problem Statement}
\label{problemStatement}
Consider a group of $N$ quadrotor UAVs, with the dynamics described by (\ref{UAVdynamics}), located in an arbitrary area within the bounded and convex inspection region $\mathcal{Q} \subset \mathbb{R}^3$. It is assumed that each UAV $i\in\mathcal{V}$, where $\mathcal{V} = \{1,\ldots,N\}$, is equipped with a gimballed camera that has the ability to rotate its field-of-view (FoV) to any direction in $\mathcal{Q}$, capturing important information about the environment. Let an object of interest, e.g. a structure, be represented by a surface from which a set of points $\mathcal{Q}_T = \{\mathbf{q}_T^l\}$, $l \in \{1,\ldots,|\mathcal{Q}_T|\}$ is extracted, where $\mathbf{q}_T^l$ is the $l^{\text{th}}$ target point on the object's boundary $\partial\mathcal{W}$, and the total number of these points is denoted by $|\mathcal{Q}_T|$. The target points that need to be inspected are already known to UAVs. The 3D inspection problem addressed in this work can be stated as follows: \textit{Find for each quadrotor UAV $i\in\mathcal{V}$ a distributed control law, such that, for any initial position $\mathbf{p}_i(0)$ of each UAV $i\in\mathcal{V}$, the multi-UAV system cooperatively inspects all target points $\mathcal{Q}_T$ on the object's boundary $\partial\mathcal{W}$, and all signals remain bounded}. The aforementioned problem can be expressed at a high level, as formulated in Problem (P1).\\[3mm]
\noindent \textbf{Problem (P1): }\textit{3D Inspection Problem}
\begin{subequations}
    \begin{flalign}
        & \quad\qquad\quad \argmin_{\boldsymbol{u}_1(t),\ldots,\boldsymbol{u}_N(t)} \mathcal{H}, \quad t\geq 0 & \label{costFunction}\\
        &\textbf{subject to: } \notag& \\
        &\dot{\mathbf{p}}_i(t) = \boldsymbol{\upsilon}_i(t),\quad \dot{\boldsymbol{\upsilon}}_i(t) = \boldsymbol{u}_i(t) & \forall i\in\mathcal{V}, \forall t\geq 0 \label{constraintDynamics}\\ 
        &\mathbf{p}_i(t)\in \mathcal{Q}-\mathcal{W}
        & \forall i\in\mathcal{V}, \forall t\geq 0 \label{constraintObject}\\
        &\mathbf{p}_i(t) \neq \mathbf{p}_j(t)         & \forall i,j\in\mathcal{V},\forall t\geq 0 \label{constraintUAVs}
    \end{flalign}
\end{subequations}

\noindent The objective is to design, for each UAV $i\in\mathcal{V}$, a control input $\boldsymbol{u}_i(t)$ that the trajectories of the multi-UAV system drive the inspection cost function $\mathcal{H}$ to the minimizer, subject to the constraints in (\ref{constraintDynamics})-(\ref{constraintUAVs}). The constraint in (\ref{constraintDynamics}) corresponds to the UAVs' dynamics introduced in (\ref{UAVdynamics}) while the constraint (\ref{constraintObject}) ensures that the position of any UAV $i\in\mathcal{V}$ for $\forall$ $t\geq 0$, belongs to the inspection region $\mathcal{Q}$ avoiding collision with the object of interest $\mathcal{W}$. Finally, the collision avoidance between agents is represented by the constraint in (\ref{constraintUAVs}).

\section{Proposed Approach}
\label{proposedApproach}
\subsection{Cost Function}

Consider a multi-UAV system consisting of $N$ quadrotor UAVs. Let $\mathcal{Q} \subset \mathbb{R}^3$ be a convex region in which the UAVs move and the position of the $i^{\text{th}}$ UAV is denoted by $\mathbf{p}_i$. The set of all UAV positions is also defined as $\mathcal{P} = \{\mathbf{p}_1,\ldots,\mathbf{p}_N\}$. The sensing quality at the point $\mathbf{q}\in\mathcal{Q}$ measured by the $i^{\text{th}}$ UAV located at $\mathbf{p}_i$ decreases proportionally with distance $\Vert \mathbf{q} - \mathbf{p}_i \Vert$. This measure can be expressed by an isotropic, strictly increasing, and convex function $f(\Vert\mathbf{q}-\mathbf{p}_i\Vert) : \mathbb{R}_{+} \rightarrow \mathbb{R}_{+}$, called the sensing unreliability function, in which the sensing quality deteriorates as large values are reached. The function $\varphi : \mathcal{Q}\rightarrow \mathbb{R}_+$ is a density function on $\mathcal{Q}$ that allocates weight to each point $\mathbf{q}\in\mathcal{Q}$ in the region, signifying the relative importance of various regions in $\mathcal{Q}$. As a result, the multi-UAV team focuses on areas with high values of $\varphi(\mathbf{q})$. Given the set $\mathcal{P}$, the optimal partition of $\mathcal{Q}$ can be obtained by constructing the set of Voronoi regions, $V = \{V_1, V_2, . . . , V_N\}$, where the positions of the UAVs are the generating points, as:
\begin{equation}
    V_i = \{\mathbf{q}\in \mathcal{Q} \text{ : } \Vert \mathbf{q}-\mathbf{p}_i\Vert \leq \Vert \mathbf{q}-\mathbf{p}_j\Vert, \mbox{ } \forall{} \mbox{ } j\neq i\}.
    \label{voronoi}
\end{equation}
If the Voronoi regions $V_i$ and $V_j$ are adjacent, that is, it holds true that $V_i \cap V_j \neq 0$, then the $i^{\text{th}}$ and $j^{\text{th}}$ are called neighbors. The neighborhood of the $i^{\text{th}}$ UAV is denoted by $\mathcal{N}_i$ comprising all neighbors of the UAV.

The inspection cost function is thereafter defined as an indicator of multi-UAV system performance as follows:
\begin{equation}
    \mathcal{H}(\mathcal{P}) = \sum_{i=1}^N\int_{V_i} f(\Vert\mathbf{q}-\mathbf{p}_i\Vert) \varphi(\mathbf{q})d\mathbf{q}.
\end{equation}
The function $\mathcal{H}$ measures how ineffective UAVs are positioned within $\mathcal{Q}$ based on its importance regions and, therefore, the multi-UAV system aims to minimize it. One way to find the minimizer of $\mathcal{H}$ is to compute its gradient concerning the UAVs' positions $\mathbf{p}_i$, which is given by, \cite{cortes2004coverage}:
\begin{equation}
    \dfrac{\partial \mathcal{H}}{\partial\mathbf{p}_i} = \int_{V_i}\dfrac{\partial f(\Vert\mathbf{q}-\mathbf{p}_i\Vert)}{\partial \mathbf{p}_i}  \varphi(\mathbf{q})d\mathbf{q} = M_{V_i}(\mathbf{p}_i - \mathbf{C}_{V_i})
    \label{partialH}
\end{equation}
where we utilize the quadratic sensing unreliability function as $f(\Vert\mathbf{q}-\mathbf{p}_i)\Vert) = \frac{1}{2}\Vert\mathbf{q}-\mathbf{p}_i\Vert^2$, and the mass $M_{V_i}$ and centroid $\mathbf{C}_{V_i}$ of the Voronoi region $V_i$ are, respectively, expressed as:
\begin{align}
    M_{V_i} = \int_{V_i}\varphi(\mathbf{q}) d\mathbf{q}, \quad 
    \mathbf{C}_{V_i} = \dfrac{1}{M_{V_i}}\int_{V_i}\mathbf{q}\varphi(\mathbf{q}) d\mathbf{q}.
    \label{massCentroid}
\end{align}
It is obvious that the partial derivative with respect to the $i^{\text{th}}$ UAV position is determined only by its own position and the positions of its Voronoi neighbors. The equilibrium points of $\mathcal{H}$ can be found when $\partial\mathcal{H}/\partial\mathbf{p}_i=0$, that is, $\mathbf{p}_i = \mathbf{C}_{V_i}$ for all $i\in\mathcal{V}$. Therefore, the multi-UAV system achieves Centroidal Voronoi Tessellation (CVT) by each UAV being positioned at the centroid of its Voronoi region.

\subsection{3D Target Points Inspection}
As mentioned above, the goal of the multi-UAV system is to inspect a set of target points $\mathcal{Q}_T$ on the surface of the object boundary $\partial \mathcal{W}$. We propose the outward projection of these points $\Bar{\mathcal{Q}}_T = \{\Bar{\mathbf{q}}_T^l\}$, $l \in\{1,\ldots,|\Bar{\mathcal{Q}}_T|\}$, as shown in Figure \ref{fig:projectedTargetPoints}, as a method of driving each UAV $i\in\mathcal{V}$ to positions around the projected points that are capable of inspecting the corresponding target points on the surface of the object. More specifically, at each projected target point $\Bar{\mathbf{q}}_T ^l$, we attach a 3D Gaussian function $\varphi_l(\mathbf{q},\Bar{\mathbf{q}}_T^l)$ centered at $\Bar{\mathbf{q}}_T ^l$, and thus the density function $\varphi(\mathbf{q})$ can be defined as follows:
\begin{equation}
    \varphi(\mathbf{q}) = 
    \sum_{l = 1}^{|\bar{\mathcal{Q}}_T|} b_l\varphi_l(\mathbf{q},\Bar{\mathbf{q}}_T^l) = 
    \sum_{l = 1}^{|\bar{\mathcal{Q}}_T|} b_l\alpha e^{-\beta \Vert \mathbf{q} - \Bar{\mathbf{q}}_T^l \Vert^2},
\end{equation}
where $\alpha,\beta > 0$, $|\bar{\mathcal{Q}}_T|$ is the total number of the projected target points, and $b_k$ is a binary variable that represents the target inspection status of $\mathbf{q}_T^l$. It is required that all target points are known to the UAVs and that a set $\mathcal{B}_i=\{b_l\}$, $\forall$ $l \in\{1,\ldots,|\mathcal{Q}_T|\}$ is stored by the UAV $i\in\mathcal{V}$ and shared with each neighbor $j\in\mathcal{N}_i$. A target point $\mathbf{q}_T^l$ is considered as inspected from the multi-UAV system if the following condition holds:
\begin{equation}
    f(\Vert \bar{\mathbf{q}}_T^l - \mathbf{p}_i \Vert) = \dfrac{1}{2}\Vert \bar{\mathbf{q}}_T^l - \mathbf{p}_i \Vert^2 \leq r \quad i\in\mathcal{V}.
    \label{inspectionStatus}
\end{equation}
The inequality in Eqn. \eqref{inspectionStatus} can be interpreted as follows. If the UAV $i\in\mathcal{V}$ is positioned within a radius $r\in\mathbb{R}_+$ from the projected target point $\Bar{\mathbf{q}}_T^l$, then it is assumed that it will acquire significant information about the target point $\mathbf{q}_T^l$, by rotating its camera FoV in that direction. Therefore, the UAV $i\in\mathcal{V}$ has inspected $\mathbf{q}_T^l$, and sets $\mathcal{B}_{i,l} = b_l = 0$. The updated target inspection status set $\mathcal{B}_{i}$ is shared with the UAV's neighbors and the density function is computed again. This procedure is repeated until all target points are inspected from the multi-UAV system.

\subsection{Object Avoidance}
During the mission of the multi-UAV system, each UAV $i\in\mathcal{V}$ moves toward the centroid $\mathbf{C}_{V_i}$ of its Voronoi region $V_i$. However, this movement generates a trajectory that may cause a collision between the UAV and the object of interest. Consequently, an object avoidance technique is adopted to maintain the UAVs' positions $\mathbf{p}_i \in \mathcal{Q}-\mathcal{W}$. As mentioned above, the object of interest is represented by a set of target points $\mathcal{Q}_T$ that all UAVs should avoid due to disastrous consequences. As a result, a repulsive function is utilized and given as follows:
\begin{equation}
    U_{o,i} = \left\{
    \begin{array}{c}
        \sum\limits_{l=1}^{|\mathcal{Q}_T|}\frac{1}{2}\epsilon\left(\frac{1}{\Vert \mathbf{p}_i - \mathbf{q}_T^l \Vert} - \frac{1}{d_o}\right)^2, \quad\mbox{if } \Vert \mathbf{p}_i - \mathbf{q}_T^l \Vert \leq d_o\\
        0 \qquad\qquad\qquad,\quad \mbox{otherwise} 
    \end{array}
    \right.
\end{equation}
where $\epsilon,d_o > 0$ are a positive gain and the safety distance, that is, the minimum allowable distance of the UAV $i\in\mathcal{V}$ from each target point $\mathbf{q}_T^l$, respectively.

\subsection{Distributed Control Design}
This section presents the design of a distributed inspection control law for each UAV $i\in\mathcal{V}$, which is governed by (\ref{UAVdynamics}). The main objective of this control law is to drive the multi-UAV system to inspect an object of interest. To achieve this goal, certain standard assumptions have been made.\\[-3mm]
\begin{assumption}
    Each quadrotor UAV is capable of computing its own Voronoi region in a distributed way.
    \label{ass:voronoiComputation}
\end{assumption}
\begin{assumption}
    Each quadrotor UAV has the ability to communicate with its Voronoi neighbors and share information.
    \label{ass:communication}
\end{assumption}

\vspace{-3mm}
For each UAV $i\in\mathcal{V}$, we design a distributed control law as follows:

\begin{equation}
    \boldsymbol{u}_i = \boldsymbol{u}_{c,i} + \boldsymbol{u}_{o,i}
    \label{controlScheme}
\end{equation}
to ensure complete inspection of the object of interest while maintaining the safety of the multi-agent system. The first term is a proportional-derivative (PD) controller that drives the agent towards centroid $C_{V_i}$ given by 

\begin{equation}
    \boldsymbol{u}_{c,i} = k_pM_{V_i}\mathbf{C}_{V_i} - (k_pM_{V_i}\mathbf{p}_i + k_d\boldsymbol{\upsilon}_i) 
    \label{controlCvi}
\end{equation}
where $k_p,k_d > 0$ are, respectively, the proportional and derivative control gains. The second term corresponds to an object avoidance controller $\boldsymbol{u}_{o,i} = - \nabla_{\mathbf{p}_i}U_{o,i}$ that ensures the multi-agent system safety given in more detailed by

\begin{align}
    \boldsymbol{u}_{o,i} &= \sum\limits_{l = 1}^{|\mathcal{Q}_T|}\mu_o c_{il}\left(\dfrac{1}{\Vert \mathbf{p}_i-\mathbf{q}_T^l \Vert} - \dfrac{1}{d_o}\right)\dfrac{ \mathbf{p}_i-\mathbf{q}_T^l}{\Vert \mathbf{p}_i-\mathbf{q}_T^l \Vert^2}
\end{align}
where $\mu_o$ is the gain of the obstacle-free term, $d_o$ is the minimum acceptable distance between the UAV $i$ and the object boundary, and the binary variable $c_{il}$, $\forall$ $i\in\mathcal{V}$, $\forall$ $l \in\{1,\ldots,|\mathcal{Q}_T|\}$  is defined as:
\begin{equation}
    c_{il} = \left\{
    \begin{array}{l}
        1, \quad \forall \mbox{ } \Vert \mathbf{p}_i - \mathbf{q}_T^l \Vert \leq d_o\\
        0, \quad \mbox{otherwise.} 
    \end{array}
    \right.
\end{equation}

We define a candidate Lyapunov function as
\begin{equation}
    \Upsilon =  k_p\mathcal{H} + \sum_{i = 1}^N \dfrac{1}{2}\boldsymbol{\upsilon}_i^T \boldsymbol{\upsilon}_i + \sum_{i=1}^N U_{o,i}.
\end{equation}
The stability of the multi-UAV system using the control scheme (\ref{controlScheme}) can be demonstrated by considering the following two cases. In the first case, if the UAV $i\in\mathcal{V}$ is outside any repulsive region, that is, $\Vert \mathbf{p}_i - \mathbf{q}_T^l \Vert \geq d_o$ for all $l \in{1,\ldots,|\mathcal{Q}_T|}$, the third term of the Lyapunov function disappears and the obstacle avoidance term of the proposed control law is removed due to all binary variables $c_{il} = 0$. Consequently, asymptotic stability can be easily proved. In the second case, if the UAV $i\in\mathcal{V}$ is within at least one repulsive region such that $\Vert \mathbf{p}_i - \mathbf{q}_T^l \Vert < d_o$, the corresponding binary variables $c_{il} = 1$, and therefore $\boldsymbol{u}_{o,i} \neq 0$. Following a similar analysis, the asymptotic stability of the multi-agent system is derived.

\vspace{5pt}
\begin{remark}
The computation of the control scheme (\ref{controlCvi}) is accomplished through the exchange of information between the $i^{\text{th}}$ UAV and its Voronoi neighbors to calculate the mass and centroid of its Voronoi region. Therefore, this control scheme is classified as distributed.
\end{remark}

\begin{remark}
    The $i^{\text{th}}$ UAV's motion tends to the centroid of its Voronoi region when the control scheme (\ref{controlCvi}) is employed. Since the centroid always lies inside the Voronoi region, and the Voronoi tessellations generate non-overlapping regions, there will be no inter-UAV collision during the mission.
\end{remark}

\vspace{5pt}

As we have presented the entire methodology, the overall process is outlined in Algorithm \ref{algorithm}. In summary, each UAV acquires the position of its neighbors to compute the centroid of its Voronoi region, which is then used in the distributed control law. Additionally, each UAV combines its target inspection status set with the sets of its corresponding neighbors, resulting in an updated density function that directs the UAV to uninspected regions.

\begin{algorithm}
    \caption{3D Inspection Algorithm}\label{alorithm}
    \begin{algorithmic}[1]
        \Require A group of UAVs $\mathcal{V} = \{1,\ldots,N\}$, with initial positions $\mathbf{p}_i(0)$, $i\in\mathcal{V}$, are located within the inspection region $\mathcal{Q}\in\mathbb{R}^3$.  Target points $\mathbf{q}_T^l\in\mathcal{Q}_T$ are known to the multi-UAV system. Each UAV $i\in\mathcal{V}$ should be able to compute its Voronoi region $V_i$, and share information with its Voronoi neighbors.
        \Ensure Complete inspection of the target points $\mathbf{q}_T^l\in\mathcal{Q}_T$.
        \While{$\mathcal{B}_i \neq \emptyset$}
            \State  \multiline{%
            \hspace{-10pt} Acquires the position $\mathbf{p}_j$ and target inspection status set $\mathcal{B}_j$, $\forall$ $j\in\mathcal{N}_i$}
            \State \hspace{-10pt} Updates the set $\mathcal{B}_i = \mathcal{B}_i \vee \mathcal{B}_j$ 
            \State \hspace{-10pt} Constructs its Voronoi region $V_i(\mathbf{p}_i,\mathbf{p}_j)$, as in (\ref{voronoi})
            \State \hspace{-10pt} Computes the centroid $\mathbf{C}_{V_i}$ of its $V_i(\mathbf{p}_i,\mathbf{p}_j)$, as in (\ref{massCentroid})
            \State  \multiline{%
            \hspace{-10pt} Computes and applies the control input $\boldsymbol{u}_i$, as in (\ref{controlScheme})}
            \State \hspace{-10pt} Update the target inspection status $\mathcal{B}_i$:
                \If{$f(\Vert \Bar{\mathbf{q}}_T^l - \mathbf{p}_i \Vert) \leq r$}
                    \State $\mathcal{B}_{i,k} = 0$, $\forall$ $l \in\{1,\ldots,|\mathcal{Q}_T|\}$
                \EndIf
        \EndWhile
    \end{algorithmic}
    \label{algorithm}
\end{algorithm}

\begin{figure*}[t]
     \begin{subfigure}[b]{0.32\textwidth}
        \includegraphics[clip, trim=2cm 11cm 9cm 5cm, scale = 0.5]{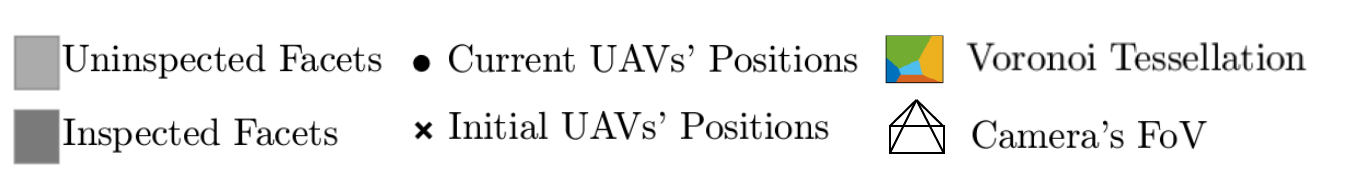}
    \end{subfigure}
    
    \centering
     \begin{subfigure}[b]{0.32\textwidth}
        \includegraphics[clip, trim=0cm 7cm 1cm 7cm, width=\textwidth]{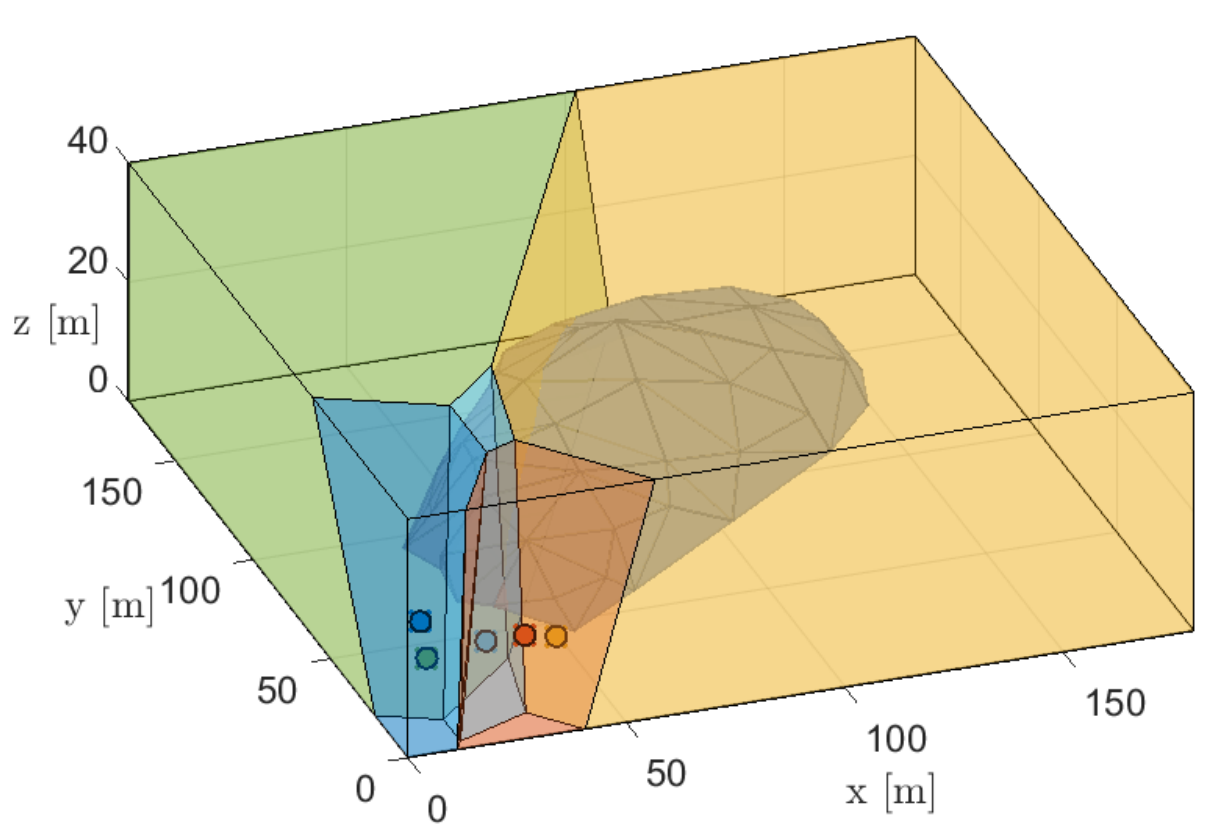}
        \caption{}
        \label{fig:1,1}
    \end{subfigure}
    \hfill
    \begin{subfigure}[b]{0.32\textwidth}
        \includegraphics[clip, trim=0cm 7cm 1cm 7cm, width=\textwidth]{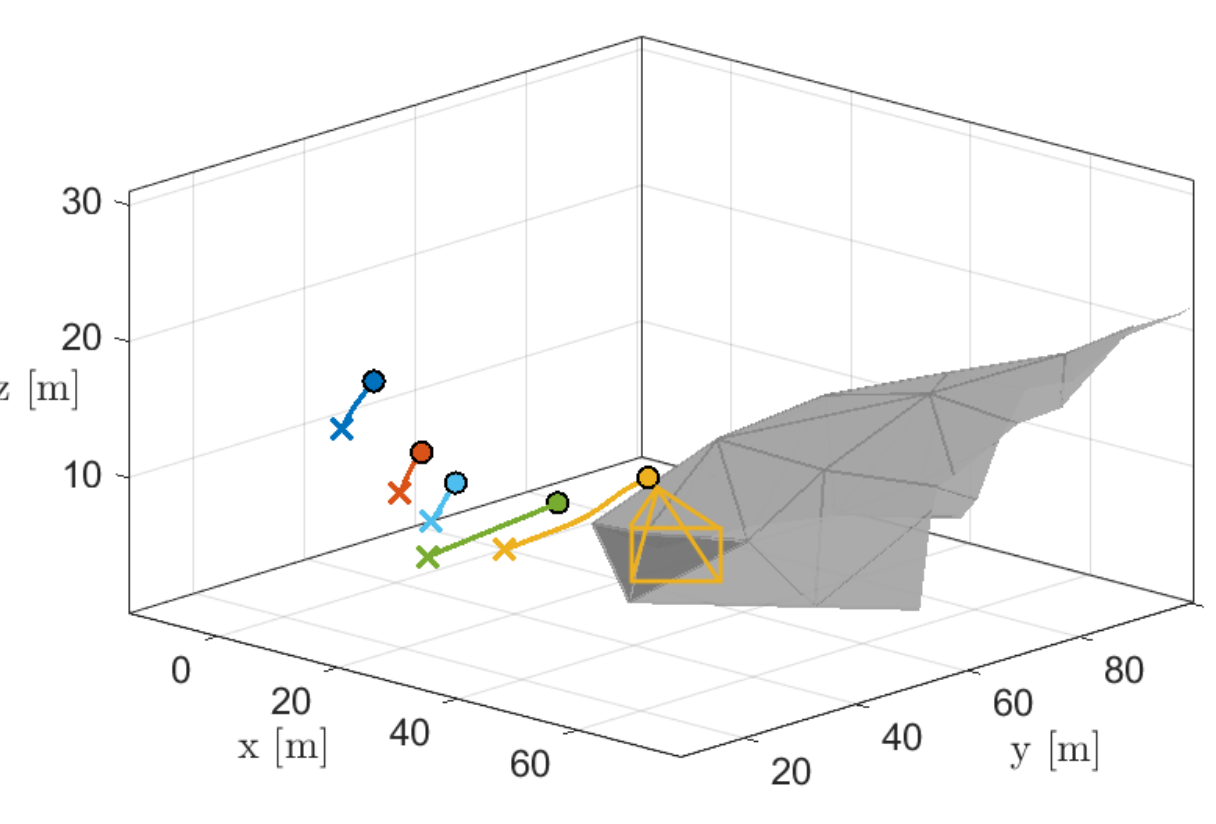}
        \caption{}
        \label{fig:1,2}
    \end{subfigure}
    \hfill
    \begin{subfigure}[b]{0.32\textwidth}
        \includegraphics[clip, trim=0cm 7cm 1cm 7cm, width=\textwidth]{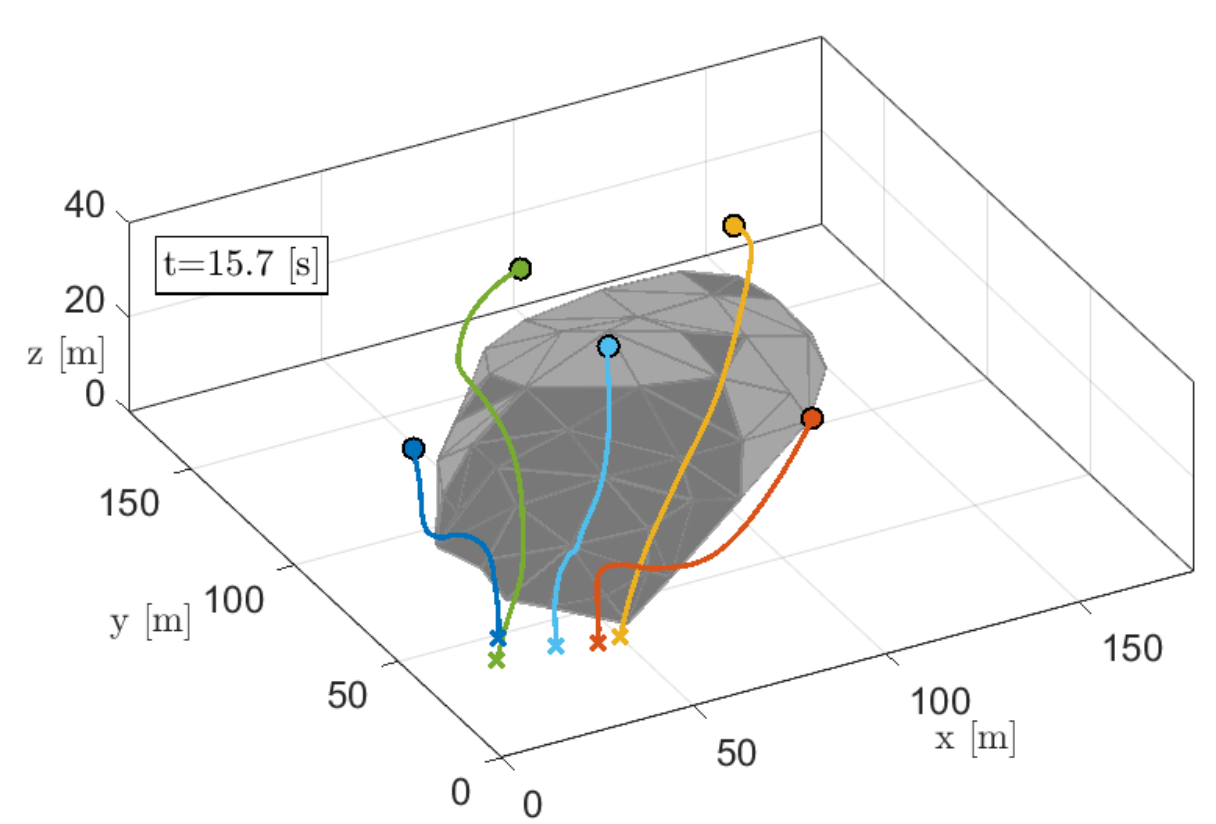}
        \caption{}
        \label{fig:1,3}
    \end{subfigure}
    
    \begin{subfigure}[b]{0.32\textwidth}
        \includegraphics[clip, trim=0cm 7cm 1cm 7cm, width=\textwidth]{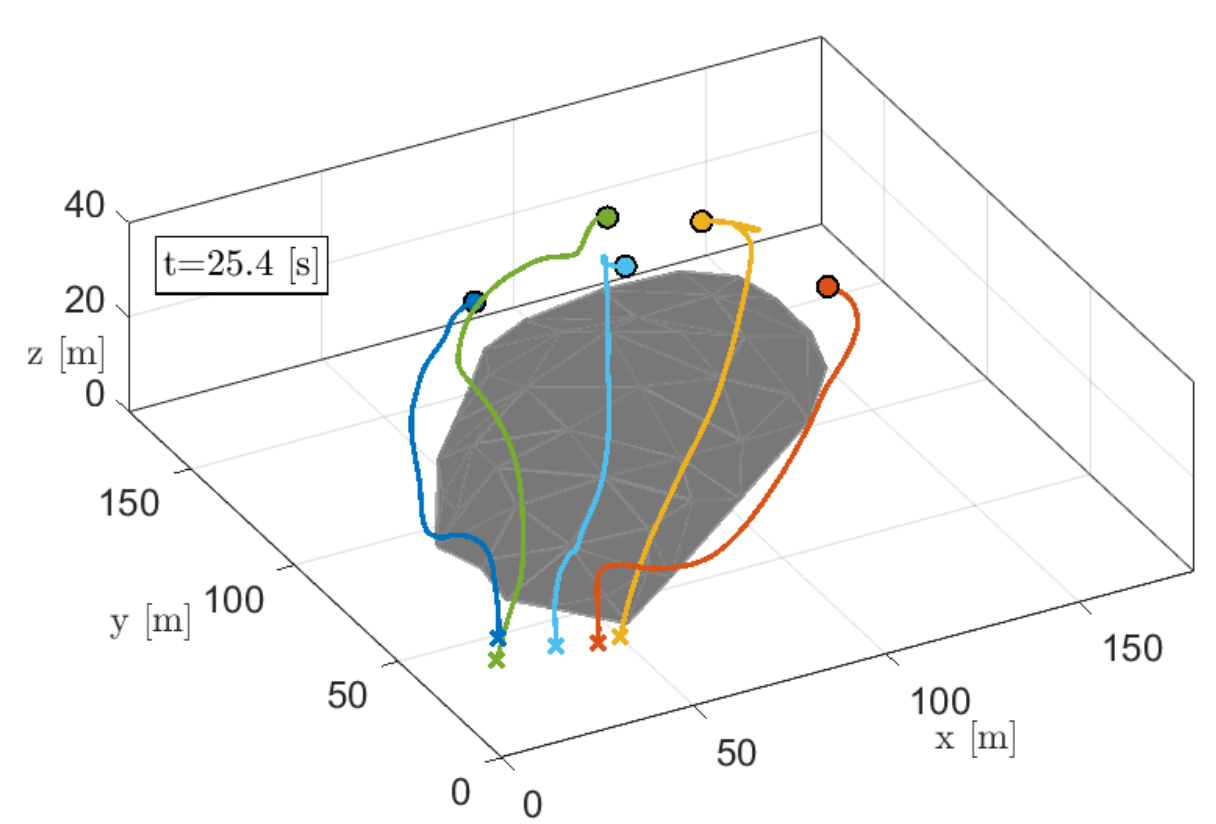}
        \caption{}
        \label{fig:2,1}
    \end{subfigure}
    \hfill
    \begin{subfigure}[b]{0.32\textwidth}
        \includegraphics[clip, trim=2cm 8cm 4cm 7cm, width=\textwidth]{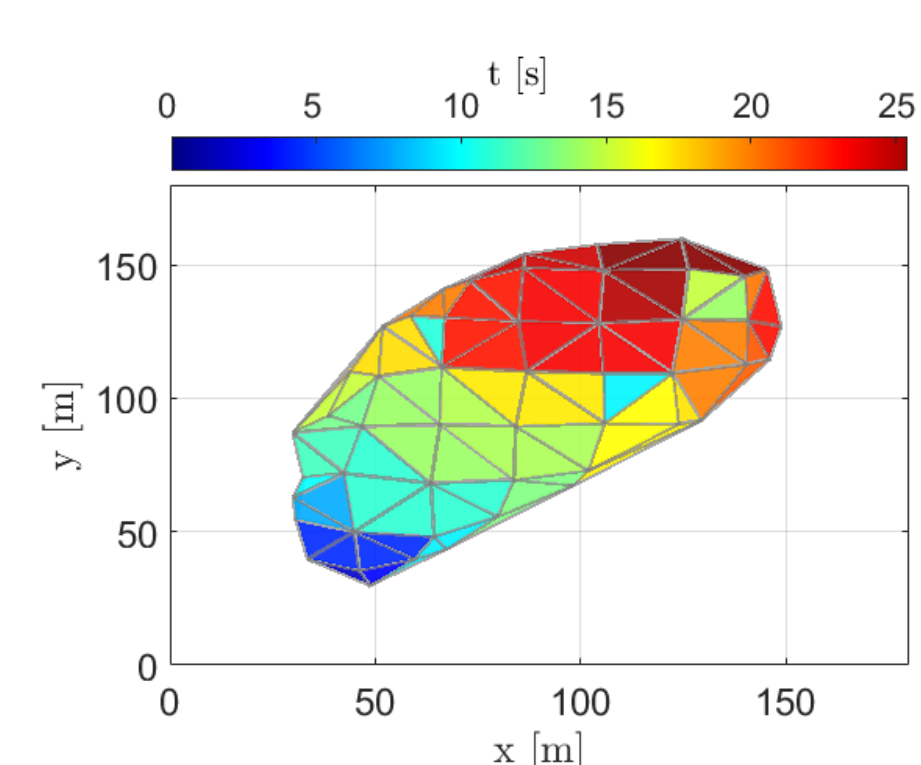}
        \caption{}
        \label{fig:2,2}
    \end{subfigure}
    \hfill
    \begin{subfigure}[b]{0.32\textwidth}
        \includegraphics[clip, trim=0cm 6cm 0cm 5cm,  width=\textwidth]{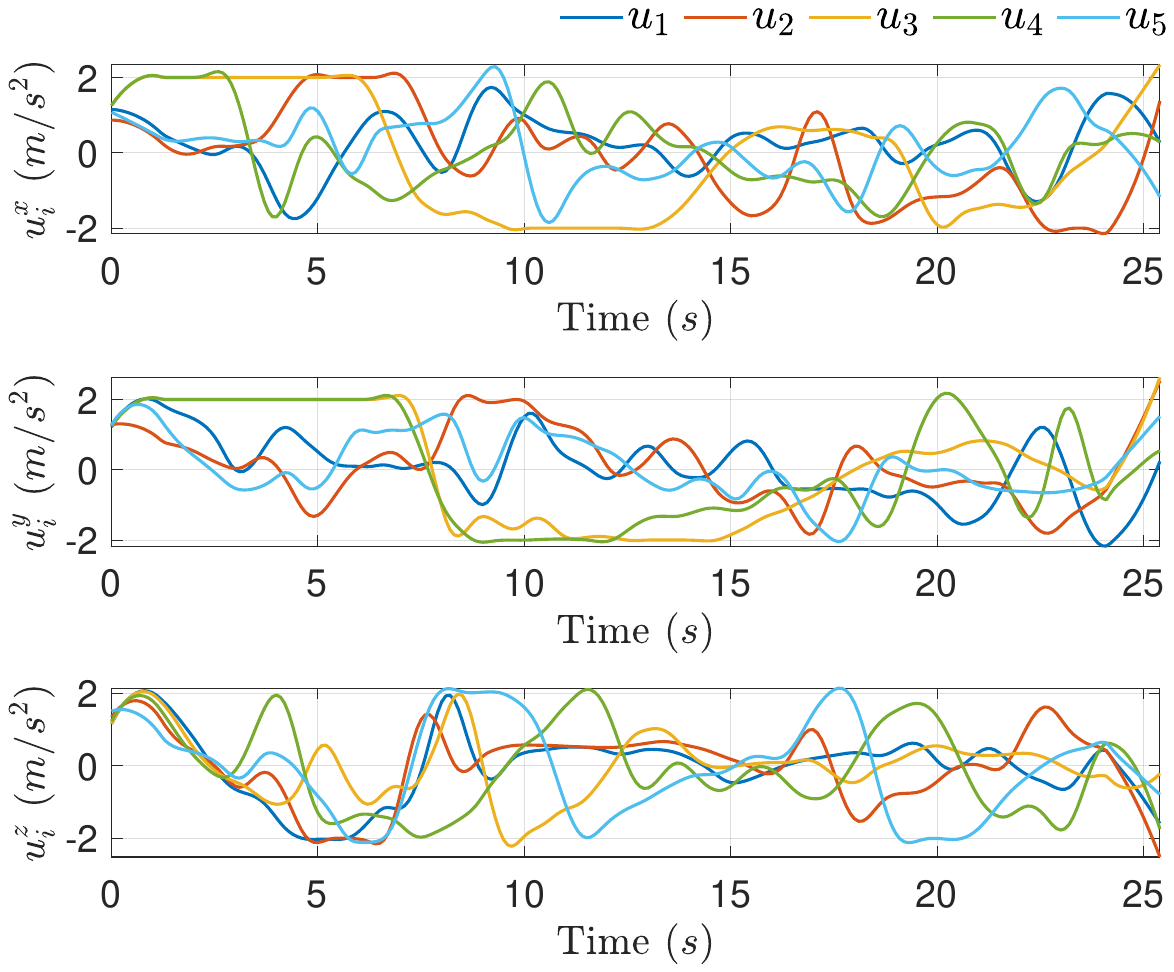}
        \caption{}
        \label{fig:2,3}
    \end{subfigure}
            
    \caption{(a) Voronoi tesselation of 
 the inspection region at $t = 0s$, (b) Rotated FoV in the direction of the target point on the object's surface $\partial\mathcal{W}$, (c) Inspection status up to $t = 15.7s$, (d) Full inspection of the object of interest, (e) Time-based inspection status, and (f) Control inputs of all UAVs}
    \label{fig:trajectoriesView}
\end{figure*}

\section{Evaluation}
\label{evaluation}
Numerous simulations have been performed to demonstrate the efficacy of the proposed methodology for the 3D inspection of an object of interest by a multi-UAV system.

\subsection{Simulation Setup}
The following numerical simulation was conducted in a MATLAB environment on a $2.8$GHz desktop computer with $16$GB of RAM. The scenario involves $5$ quadrotor UAVs, governed by dynamics (\ref{UAVdynamics}), which are initially located inside a cuboid inspection region of dimensions $180$m $\times$ $180$m $\times$ $40$m. The objective of the multi-UAV system is to inspect an object of interest of dimensions $156$m $\times$ $78$m $\times$ $26$m. More specifically, a set of $132$ target points, being part of the object's surface, have to be inspected by the multi-UAV system. To achieve this goal, density functions are used with the parameters $\alpha = 1$ and $\beta = 0.0075$ while the sensing range of each UAV is chosen as $r = 10$m. The control gains are selected as $k_p = 0.32$, $k_d = 0.86$, and $\mu_o = 1000$ for a safety region $d_o = 12$m.

\subsection{Results}
A group of five UAVs is initially located within the inspection region at the random positions $\mathbf{p}_1^0 = [10,20,15]^T$m, $\mathbf{p}_2^0 = [30,9,14]^T$m, $\mathbf{p}_3^0 = [40,17,10]^T$m, $\mathbf{p}_4^0 = [15,30,5]^T$m, and $\mathbf{p}_5^0 = [25,20,10]^T$m, respectively. Each UAV initializes its target inspection status set with ones, as all target points are uninspected. By acquiring the positions of its neighbors, UAV $i$ constructs its Voronoi region. The initial Voronoi tessellation of the inspection region is illustrated in Fig. \ref{fig:1,1}, with the generating points being the initial positions of the UAVs. All Voronoi cells are colored with the corresponding UAV's color. Afterward, each UAV calculates the centroid $C_{V_i}$ of its corresponding Voronoi cell and safely moves towards it. As the UAV $i$ moves in the direction of $C_{V_i}$, the projected target point $\bar{\mathbf{q}}_T^l$ enters its sensing region. Then, the camera's FoV is automatically rotated pointing to the corresponding target point $\mathbf{q}_T^l$ on the object's surface, as depicted in Fig. \ref{fig:1,2}. As a result, this target point is captured with satisfactory quality and is marked as inspected by $\mathcal{B}_{i,l} = 0$. However, a facet, which consists of four target points (the facet vertices and center), is considered inspected when all these target points are inspected, and only then it is colored dark gray. Fig. \ref{fig:1,3} illustrates the trajectories of UAVs and the inspected/uninspected facets, at the time instant $t = 15.7s$. A safe 3D inspection of the object of interest requires $25.4s$ to be completed, and the resulting trajectories are presented in Fig. \ref{fig:2,1}. More precisely, a time-based facet inspection status is provided in Fig. \ref{fig:2,2} which shows at which time instant a facet was inspected. The order of facet inspection is not unique and depends on the initial positions of the UAVs. Finally, in the last Fig. \ref{fig:2,3} are shown the control inputs in all three dimensions (x,y,z) of all UAVs during the mission, resulting in collision-free trajectories that completely inspect the object of interest.

\section{Conclusion}
\label{conclusion}
In this work, we have proposed a methodology that solves the problem of 3D object inspection using a multi-UAV system. The objective function was minimized by designing a distributed control law that generates safe trajectories for the multi-UAV system, achieving complete object inspection. Finally, the effectiveness of the proposed approach was demonstrated through simulations. Future work may consider trajectory optimization and real-world experiments.

\section*{Acknowledgment}
This work was undertaken as part of the GLIMPSE project EXCELLENCE/0421/0586 which is co-financed by the European Regional Development Fund and the Republic of Cyprus through the Research and Innovation Foundation’s RESTART 2016-2020 Programme for Research, Technological Development and Innovation and supported by the European Union’s Horizon 2020 research and innovation programme under grant agreement No 739551 (KIOS CoE), and from the Government of the Republic of Cyprus through the Cyprus Deputy Ministry of Research, Innovation and Digital Policy.\\


\bibliography{IEEEabrv,main}

\end{document}